\pdfoutput=1

\documentclass[11pt, amsart]{article}

\usepackage{EMNLP2022}

\usepackage{times}
\usepackage{latexsym}

\usepackage[T1]{fontenc}

\usepackage[utf8]{inputenc}

\usepackage{microtype}

\usepackage{inconsolata}

\usepackage{booktabs}

\usepackage{tikz}
\def\checkmark{\tikz\fill[scale=0.4](0,.35) -- (.25,0) -- (1,.7) -- (.25,.15) -- cycle;} 

\usepackage{xcolor}

\usepackage{amsfonts}

\usepackage{subcaption}

\title{Tackling Low-Resourced Sign Language Translation: UPC at WMT-SLT 22}

\author{Laia Tarrés$^{1,2}$\thanks{ $^*$Equal contribution}
\and
Gerard I. Gállego$^{1, \ast}$
\and
Xavier Gir{\'o}-i-Nieto$^{1}$ 
\and
Jordi Torres$^{2}$
\\
$^1$\emph{Universitat Polit\`{e}cnica de Catalunya, Spain} \quad $^2$\emph{Barcelona Supercomputing Center, Spain}
}

\begin{document}
\maketitle
\begin{abstract}
This paper describes the system developed at the Universitat Politècnica de Catalunya for the Workshop on Machine Translation 2022 Sign Language Translation Task, in particular, for the sign-to-text direction. 
We use a Transformer model implemented with the Fairseq modeling toolkit. We have experimented with the vocabulary size, data augmentation techniques and pretraining the model with the PHOENIX-14T dataset. 
Our system obtains $0.50$ BLEU score for the test set, improving the organizers' baseline by $0.38$ BLEU. We remark the poor results for both the baseline and our system, and thus, the unreliability of our findings.
\end{abstract}

\section{Introduction}
The submission of the Universitat Politècnica de Catalunya (UPC) to the WMT22 Sign Language Translation (SLT) Task experimented with vocabulary size, data augmentation techniques and a pretrained system with the PHOENIX-14T dataset~\cite{Phoenix2_dataset}. Up to the author's knowledge, our implementation is the first to build on Fairseq, a popular modeling toolkit by Meta AI~\cite{Fairseq}.

SLT is a highly complex task because sign language understanding requires a very precise estimation of the signer pose, especially, of its hands. In addition, sign languages have grammatical structures different from spoken languages, which prevents an easy knowledge transfer.
Sign languages are represented in continuous and high-dimensional spaces, while the transcribed version of spoken languages are represented by discrete tokens of well-defined vocabularies.
Moreover, the few available sign language datasets can be considered very low-resourced, as they typically contain less than one hundred thousand sentences~\cite{low_resource_languages}. In particular, the total number of sentences from the two datasets provided in WMT-SLT22 is 17k.

We focus on the sign language translation task of WMT-SLT22 which requires participants to predict the translation in spoken language (written) from a sign language video. Specifically, it consists of translation from Swiss German Sign Language (DSGS) videos to German (DE) text.

The organizers of WMT 2022 SLT track propose a Transformer baseline~\cite{mueller2022sign-sockeye-baselines} that achieves a very low BLEU~\cite{bleu} score,\footnote{For simplicity, we refer to BLEU-4 score as BLEU score.} which indicates a very poor translation quality. 
The training data provided consists of two datasets: FocusNews~\cite{focusnews} and SRF~\cite{srf}. The first one contains 197 episodes in DSGS that have an average length of 5 minutes, which amount for a total duration of 19 hours.
The second dataset contains 29 videos from live sign language interpretation, that have an average of 30 minutes length, and totals a duration of 16 hours. The paired subtitles are given in Standard German from Switzerland, which is a dialect of German. 

The organizers provide keypoints extracted with two off-the-shelf human pose estimators: OpenPose~\cite{OpenPose} and MediaPipe~\cite{MediaPipe}. 

\section{Baseline system}
\label{sec:baselines}
For the sake of self-containment, we briefly define and discuss the baseline model proposed by the organizers for this task~\cite{mueller2022sign-sockeye-baselines}. This solution is built on sockeye~\cite{sockeye}, with results for the official development (dev) partition.

Regarding the model implementation, they use a Transformer encoder-decoder, with a symmetrical number of layers for the encoder and decoder. The architecture has 6 layers, 8 heads, 2048 neurons in the feed forward network layers, and an embedding dimension of 512.

The input data are 2D OpenPose keypoints, concatenating the hands and body landmarks. This results in an array $X = x_0,\dots, x_T$ where $x_t \in \mathbb{R}^{2K}$, $K$ is the number of keypoints selected, and $T$ the number of video frames.

Three baseline results were provided by the WMT-SLT22 organizers, shown in Table~\ref{tab:bleu4_organizers}. The first two scores correspond to the models trained individually with each of the two benchmark datasets. The last result is from a model trained on both datasets, which obtains the lowest performance. We hypothesize this might be due to a domain shift between the FocusNews and SRF datasets.
These BLEU scores are extremely low, as already noted by~\citet{mueller2022sign-sockeye-baselines}. As a comparison, the SLT state-of-the-art BLEU score for PHOENIX-14T dataset~\cite{Phoenix2_dataset} is 25.59~\cite{stochastic_transformers_phoenix} and for the How2Sign dataset~\cite{How2Sign_dataset} is 1.25~\cite{duarte2022sign}.

\begin{table}[ht]
    \centering
    \begin{tabular}{lcccc}
    \toprule
    \textbf{Train dataset}        &\textbf{BLEU}\\
    \midrule
    FocusNews           & 0.216 \\
    SRF                 & \textbf{0.589} \\
    FocusNews + SRF     & 0.157\\
    \bottomrule
    \end{tabular}
    \caption{\label{tab:bleu4_organizers} Results provided by the organizers for the official dev partition, which contains FocusNews + SRF samples. Scores on test partition, which also contains FocusNews + SRF samples, were not released by the organizers.}
    \vspace{-0.5cm}
\end{table}

\section{Method}
Our submission also adopts a Transformer architecture, which we implement with the Fairseq sequence modeling toolkit~\cite{Fairseq}. Up to the author's knowledge, this is the first time that Fairseq is used for sign language video understanding. We publish our source code \footnote{https://github.com/mt-upc/fairseq/tree/wmt-slt22}, offering the SLT community a novel tool widely used in machine translation for spoken languages. On top of this, we experiment with the vocabulary size, data augmentation techniques and pretraining the system with PHOENIX-14T dataset. Details are described in this section.

\subsection{Preprocessing steps}
The WMT-SLT22 organizers provide both OpenPose and MediaPipe keypoints from the body pose estimators. We choose MediaPipe poses because they provide 3D coordinates $(x, y, z)$ normalized between $[0,1]$. Moreover, based on our experience, OpenPose is more prone to errors, like detecting several people in videos when there is only one signer in the recording.

Although MediaPipe poses are available from WMT-SLT22, we re-extract them with \textit{pose-format}~\cite{pose_format_helper}. This library defines a standardized way of storing poses, and provides different functionalities to work with them. After the extraction, we obtain an array with the same shape as described in Section~\ref{sec:baselines}.

While the video recordings in the SRF dataset have a rate of 25 frames per second (fps), the FocusNews dataset present a frame rates of either 25, 30 or 50 fps. We perform cubic interpolation for the extracted poses, to achieve a unified frame rate of 25fps, using the \textit{interpolate\_fps} function from \textit{pose-format}.

We build an independent vocabulary with the training split of each dataset. 
For the SRF dataset, organizers provide parallel and monolingual data. The latter contains all German subtitles, including much more sentences than the former. Therefore, we choose to build our SRF vocabulary from the monolingual data. Note that this data does not have paired (or parallel) videos, so it can not be used for training the model.

\subsection{Architecture}
\label{sec:arch}
We build a smaller Transformer architecture than the WMT-SLT22 baseline, since we observed signs of overfitting when checking the training losses. With the baseline architecture the system was simply generating the most frequent words from the training set. Therefore, we concluded that training a smaller model would hinder the overfitting and might improve the results. 

In particular, our Transformer model has a symmetrical structure for the encoder and decoder, with 3 layers, 4 heads, 1024 neurons in the feed forward network layers, and an embedding dimension of 256.

\subsection{Data augmentation}
\label{sec:da}
The performance of deep learning models depends on the quality, quantity, and domain of training data, however datasets that provide all qualities needed for models are often not available. To diminish the consequences of the data scarcity, a common practice is to apply data augmentation techniques. This approach is really useful to improve the performance of models, and makes them more robust to slight changes in the input data.

\begin{figure}[ht]
    \vspace{-0.4cm}
    \centering
    \begin{subfigure}[b]{0.23\textwidth}
        \includegraphics[width=\textwidth]{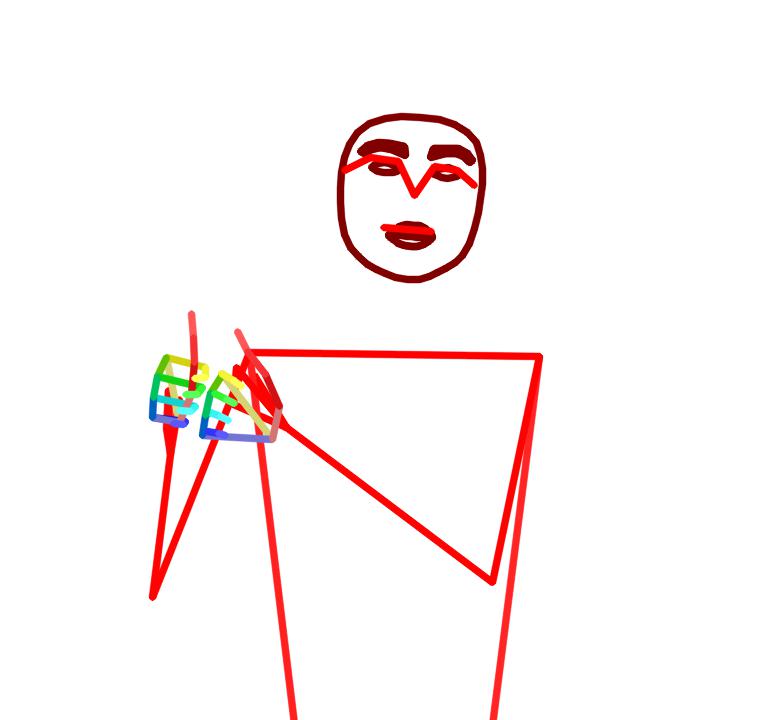}
        \caption{Original}
    \end{subfigure}
    \begin{subfigure}[b]{0.23\textwidth}
        \includegraphics[width=\textwidth]{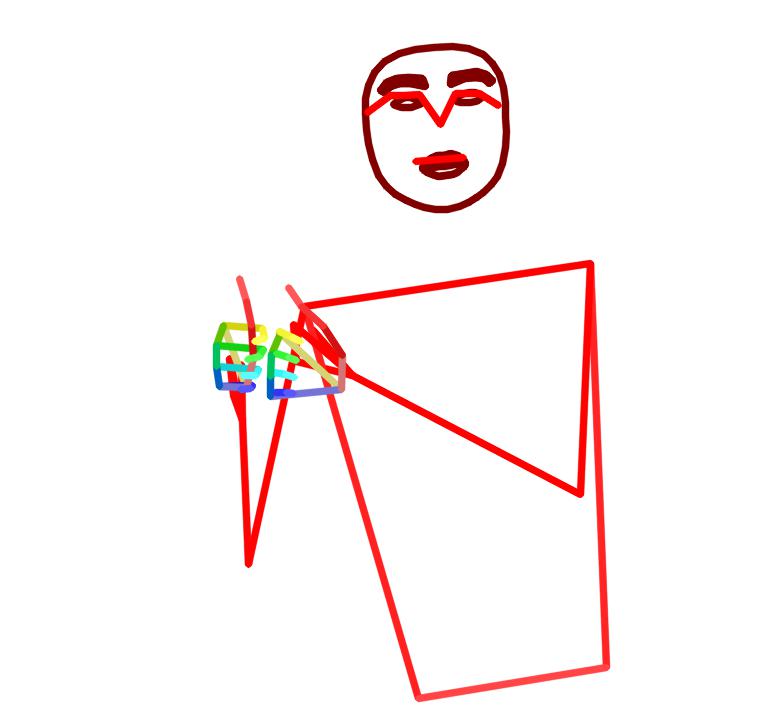}
        \caption{Rotation}
    \end{subfigure}
    \vskip\baselineskip
    \begin{subfigure}[b]{0.23\textwidth}
        \includegraphics[width=\textwidth]{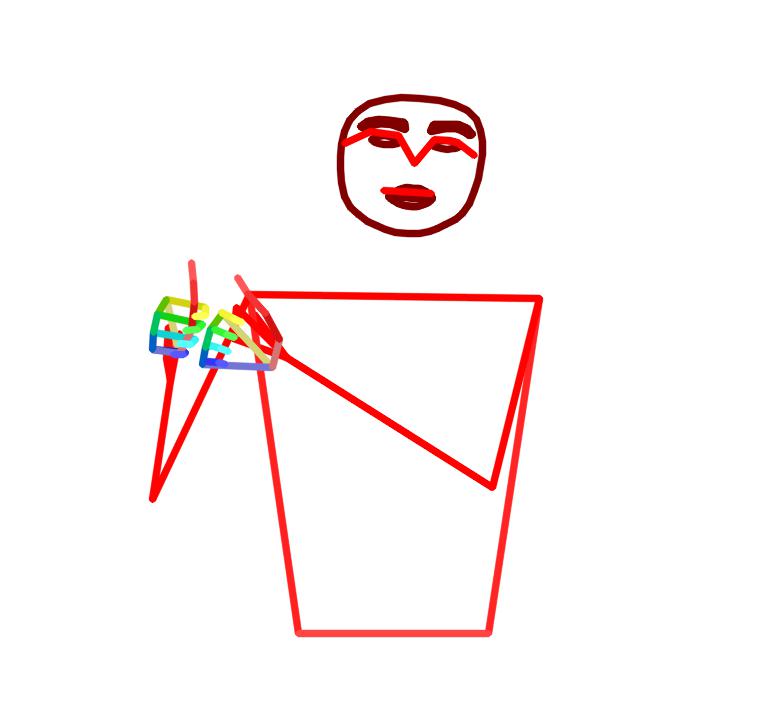}
        \caption{Scaling}
    \end{subfigure}
    \begin{subfigure}[b]{0.23\textwidth}
        \includegraphics[width=\textwidth]{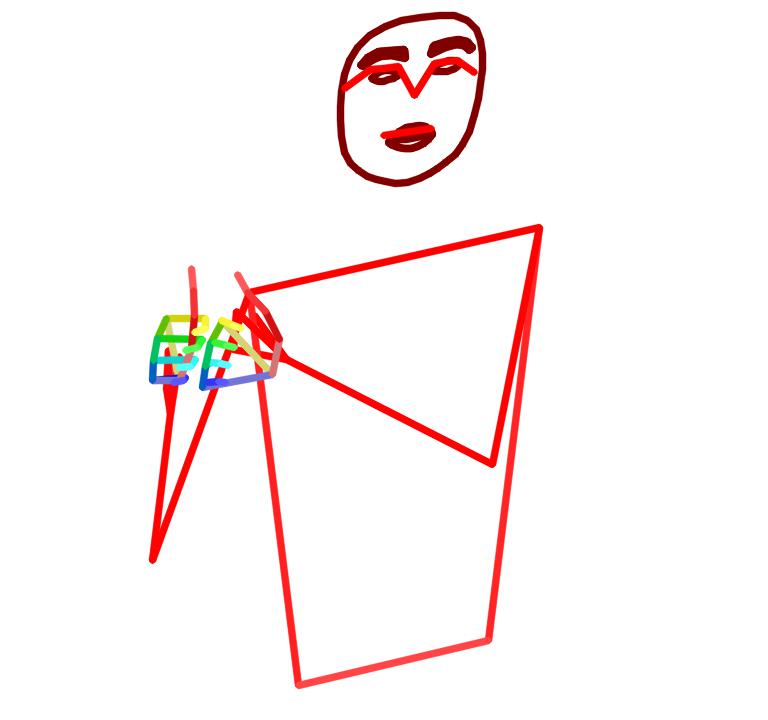}
        \caption{Shear}
    \end{subfigure}
    \caption{Data augmentation transformations.}
    \label{fig:data_augm}
    \vspace{-0.5cm}
\end{figure}

We made use of the \textit{augment2d} function from the \textit{pose-format} library, which allows applying various transformations directly to the keypoints, such as random rotation, shear effect, and scaling. Specifically, the rotation angle in radians, the shear factor and the scaling factor we apply are obtained by sampling from a normal distribution with zero mean and standard deviation of $0.2$. Some examples of augmented poses are shown in Figure~\ref{fig:data_augm}.

\subsection{Vocabulary size}
 \label{sec:vo_sz}
State-of-the-art Machine Translation systems use subword dictionaries instead of word-level vocabularies~\cite{sota_mt_news,sota_mt_multi}.
These dictionaries are built by decomposing words into smaller pieces based on their frequency~\cite{bpe_tokenizer}. Analogously to the baseline approach, we use SentencePiece tokenizer to obtain the subword vocabulary~\cite{sentencepiece}.

The vocabulary size is a hyperparameter that, in practice, is either chosen arbitrarily or via trial-and-error~\cite{choosing_vocab_size}. However, it has been studied that using a greater vocabulary size might help in reducing the class imbalance present in the training dataset~\cite{greater_vocab_size}. The baseline used a vocabulary size of 1000 subwords, and we experiment with 2000 and 4000. Our goal was to detect whether downsampling a vocabulary of ~20k unique words, for FocusNews dataset, to 1k subwords may be oversimplifying the problem. 

\subsection{Pretraining with PHOENIX14-T}
\label{sec:pt}
We also explore transfer learning to overcome the data scarcity problem.
Given that the scope of WMT22 is on Swiss German Sign Language, we chose the PHOENIX14-T dataset~\cite{Phoenix2_dataset}. For the three datasets, the target language is either spoken (written) German or Standard German from Switzerland. However, PHOENIX14-T presents an important domain shift with respect to the WMT-SLT22 datasets, since it is limited to live interpretation of weather forecast on broadcast TV.

We pretrained our model with the PHOENIX14-T dataset. In order to implement the transfer learning pipeline, we built a vocabulary by merging the training data from the three available datasets: FocusNews, SRF and PHOENIX14-T. 

\begin{table*}[ht]
    \centering
    \begin{tabular}{llccccc}
    \toprule
    \textbf{ID} & \textbf{System}   & \textbf{Vocab.}   & \textbf{Data} & \textbf{Pretrain} &\textbf{BLEU}  &\textbf{BLEU} \\
                &                   & \textbf{size}     &\textbf{Augm.} &                   &\textbf{(dev)} &\textbf{(test)} \\
    \midrule
    1 & Baseline~\cite{mueller2022sign-sockeye-baselines} & 1k & & & $0.22$ & - \\
    \midrule
    2 & Our Baseline (\S\ref{sec:arch}) & 1k    &               &               & $0.47$            & $\textbf{0.50}$    \\
    3 & 2 + 2k subwords             & 2k    &               &               & $0.47$            & -                 \\
    4 & 2 + 4k subwords             & 4k    &               &               & $0.62$            & -                 \\
    \midrule
    5 & 4 + data augmentation       & 4k    & \checkmark    &               & $0.51$            & -                 \\
    6 & 4 + pretrain w/ Phoenix     & 4k    &               & \checkmark    & $\textbf{0.64}$   & $0.41$             \\
    7 & 6 + data augmentation       & 4k    & \checkmark    & \checkmark    & $0.48$            & -                 \\
    \midrule
    8 & 6 + checkpoint average      & 4k    &               & \checkmark    & $0.57$            & $0.35$             \\
    \bottomrule
    \end{tabular}
    \caption{\label{tab:bleu4_us_fn} Results of models trained with the FocusNews dataset. \textit{BLEU (dev)} corresponds to the results obtained in the challenge dev set, and \textit{BLEU (test)} to the results extracted by the organizers using the official test partition~\cite{wmt_slt_preliminary_findings}. In bold are the best results for each partition.}
\end{table*}

\begin{table*}[ht]
    \centering
    \begin{tabular}{clccccc}
    \toprule
    \textbf{ID} & \textbf{System}   & \textbf{Vocab.}   & \textbf{Data} & \textbf{Pretrain} &\textbf{BLEU}  &\textbf{BLEU} \\
                &                   & \textbf{size}     &\textbf{Augm.} &                   &\textbf{(dev)} &\textbf{(test)} \\
    \midrule
    1 & Baseline~\cite{mueller2022sign-sockeye-baselines} & 1k & & & $0.59$ & $0.12$ \\
    \midrule
    2 & Baseline (\S\ref{sec:arch}) & 1k    &               &               & $0.64$            & -                 \\
    3 & 2 + 2k subwords             & 2k    &               &               & $\textbf{0.69}$   & $\textbf{0.28}$    \\
    4 & 2 + 4k subwords             & 4k    &               &               & $0.63$            & $\textbf{0.28}$    \\
    \midrule
    5 & 3 + checkpoint average      & 2k    &               &               & $0.60$            & $0.24$    \\
    \bottomrule
    \end{tabular}
    \caption{\label{tab:bleu4_us_srf} Results of models trained with the SRF dataset. \textit{BLEU (dev)} corresponds to the results obtained in the challenge dev set, and \textit{BLEU (test)} to the results extracted by the organizers using the official test partition~\cite{wmt_slt_preliminary_findings}. In bold are the best results for each partition.}
\end{table*}

\subsection{Checkpoint Averaging}
\label{sec:avg}
We choose the best-performing models as those with the best BLEU dev scores. However, in the best-performing cases, and as the final step in our trainings, we average the weights of the 3 best model checkpoints for each run. 
This methodology, which was firstly introduced by~\cite{attention_is_all_you_need}, proved to be a useful and easy to implement technique to generate more robust predictions on Transformers~\cite{training_tips_transformer}, and has been widely used in the Machine Translation field.

\section{Results}
We train our systems with FocusNews and SRF. For both datasets, we provide results for the dev set, which contains recordings from FocusNews and SRF. We choose the 6 best-performing models for dev to submit to the official challenge submission.

Table~\ref{tab:bleu4_us_fn} shows the results of models trained with FocusNews. We notice an improvement in the performance of our baseline implemented in Fairseq with respect to the one from the organizers (1-2). We then analyze the effect of the vocabulary size based on the BLEU obtained for the dev set. We notice that using 4000 subwords provides the best results in terms of vocabulary size (2-4). Therefore, we choose this configuration to experiment with pretraining and data augmentation. For this set of experiments, we see a slight improvement when fine-tuning a network that has been pretrained with the PHOENIX14-T dataset (6). However, adding data augmentation (5) or a combination of both data augmentation and pretraining (7) does not improve the results. Checkpoint averaging does not bring an improvement either (8). Surprisingly, we cannot extract the same conclusions from the test results. After we received the preliminary findings from the organizers~\cite{wmt_slt_preliminary_findings}, we found that the best-performing model for dev (6) was not the best for test, but the simplest one (2).

Results of models trained with SRF are presented in Table~\ref{tab:bleu4_us_srf}. Similarly to FocusNews, our proposed baseline improves the organizers' in this dataset (1-2). However, in this case, the optimal number of subwords is 2000, with a slight improvement over other vocabulary sizes (2-4). Similar to the FocusNews case, we observe that checkpoint averaging does not improve results. Due to technical issues and time limitations, experiments with SRF are limited to analyzing the vocabulary size, hence the best experiment with FocusNews could not be replicated using SRF. Although results for dev are better for models trained with SRF, results for test show a poorer performance than FocusNews models.

We optimized our systems to obtain the best BLEU metric, without taking other metrics into consideration. However, organizers also compute chrF++~\cite{chrf++} and BLEURT~\cite{bleurt} metrics~\cite{wmt_slt_preliminary_findings}. We find that the BLEURT score shows a similar performance than BLEU. However, for the chrF++ metric, which correlates better with respect to human relative rankings, our models score lower compared to other submissions.

We provide some examples of the sentences generated by our best-performing model in Table~\ref{tab:examples}. We see that the translations are poor and lack correlation with the video, which relates to the poor overall performance in the BLEU metric.

\begin{table}[ht]
    \centering
    \begin{tabular}{lp{0.35\textwidth}}
    \toprule
    Ref.:  & \textit{Letztes Jahr haben viele Gehörlosen- Medien über die erste Gehörlosen Universität in Europa in Bad Kreuznach in Deutschland berichtet.} \\
    Pred.: & \textit{Am letzten Samstag, 22. Mai, in der Schweiz, in der Schweiz, in der Schweiz, GSC Aarau, GSC Aarau.} \\
    \midrule
    Ref.:  &  \textit{Dazu sind 4 Politiker eingeladen, die über für Behinderte wichtige Themen diskutieren werden, wie zum Beispiel TV-Untertitel, UNO Konvention für Behinderte usw.}\\
    Pred.: & \textit{Das Ziel ist es, dass es, dass die nempflichkeit für die nempflichkeitssetzen kann.}\\
    \midrule
    Ref.:  &  \textit{Der Deutsche Fernsehsender ZDF bietet Filme im Internet mit Untertitel an, sofern der Film vorher im Fernseher mit Untertitel ausgestrahlt wurde.}\\
    Pred.: & \textit{Der Schweizerische Gehörlosenbund SGB-FSS organisiert mit dem Schweizerischen Gehörlosenbund SGB-FSS, der Gehörlosen Sportverband der Gehörlosen Sportverband der Gehörl osen Sportverband der Gehörlosen Sportverband der Gehörlosen Sportverband der Gehörlosen Sportverband.}\\
    \bottomrule
    \end{tabular}
    \caption{\label{tab:examples} Example reference and predictions from our best-performing model for the official dev partition.}
\end{table}

\section{Discussion and Conclusion}

We proposed a pipeline to tackle the Sign Language Translation Task for the newly released datasets: Focusnews and SRF. Our fresh implementation with Fairseq slightly improved the baseline provided by the organizers.

Our findings showed that when training with FocusNews, our baseline system has the best performance for test. The changes in vocabulary size did not affect the test performance.
Furthermore, we showed that using checkpoint averaging does not help for this task.
In all cases, we still think that the results are extremely low, which indicate a really poor translation, and there is potential unreliability of the findings due to the close to 0 BLEU score.

We consider the results we obtained can be further improved, so we leave some experiments for future work.
Firstly, we believe that a joint training from the two provided datasets could boost the performance of the models by bridging the domain gap between these datasets.
Secondly, we did not see any improvement by pretraining the models with PHOENIX14-T dataset. However, we think that solving the WMT-SLT22 task must require some sort of transfer learning from a pretrained model.

\section*{Limitations}
As stated by the organizers, results are still poor. When inspecting the predictions, it seems evident that the model is learning the most frequent words in the vocabulary, thus failing to provide meaningful predictions from the video.
We consider that this is due to the high complexity of the task paired with a lack of data.

\begin{table}[ht]
    \centering
    \begin{tabular}{lccccc}
    \toprule
                        &\textbf{Length}
                        &\textbf{Words}
                        
                        &\textbf{Ratio}\\
    \midrule
    FocusNews           &19 h   &21 k       & 0.90  \\
    SRF                 &16 h  &19 k        & 0.84  \\
    \midrule
    How2Sign            &79 h   &16 k       & 4.93  \\
    PHOENIX14-T         &11 h   &3 k        & 3.67  \\
    \bottomrule
    \end{tabular}
    \caption{\label{tab:vocabs} Comparison between SLT datasets based on the duration of the videos (in hours) and number of unique words (in thousands) in the vocabulary. The Ratio column provides an indication of the difficulty of solving the SLT task for each dataset.}
    \vspace{-0.6cm}
\end{table}

As shown in Table~\ref{tab:vocabs}, the ratio between the training data and vocabulary size is much lower compared to other SLT datasets such as How2Sign and PHOENIX14-T. We take these results as an indication of the complexity of the datasets. We hypothesize that the low BLEU scores reported in the baseline, may be caused by the low ratio between video hours per unique words in the vocabulary, hence the dataset might be too complex. Therefore, we decide to experiment with data augmentation since it artificially improves the amount of training samples.

Our experiments with the SRF dataset, have been computationally expensive. Due to technical details, we have to read full sequences of around 30 minutes every time we load a sample. Processing them, even with the dimensional reduction provided by pose estimators, has been a challenge for the machines of our academic lab. A slightly better set of results might have been produced, but we still think it would not have made a significant difference.

In addition, we present the input poses as a sequence of one-dimensional arrays with the XYZ coordinates. We think that this may not be the optimal way of processing the graph-like structure from poses. Using graph neural networks to preprocess input poses~\cite{stgcn2018aaai,mmskeleton2019,bull2020automatic,skeleton_aware_slr} might be an interesting approach to improve SLT results.

We also lacked the time to experiment with other features, such as processing the RGB videos with a convolutional network~\cite{joze2018msasl,li2020wlasl,albanie2020bsl}.
We tried extracting i3d features fine-tuned on PHOENIX14-T, but the output features contained an excessive number of 0's, and we never run a proper experiment with this setup. This might happen because the visual appearance of the videos is too different between PHOENIX14-T and the WMT-SLT22 datasets, specifically due to the spatial segmentation of the signer in the frames provided in these datasets.

\section*{Acknowledgements}
This research was partially supported by research grant Adavoice PID2019-107579RB-I00 / AEI / 10.13039/501100011033, research grants PRE2020-094223, PID2021-126248OB-I00 and PID2019-107255GB-C21. 

\bibliography{anthology,emnlp2022}

\begin{thebibliography}{32}
\expandafter\ifx\csname natexlab\endcsname\relax\def\natexlab#1{#1}\fi

\bibitem[{Albanie et~al.(2020)Albanie, Varol, Momeni, Afouras, Chung, Fox, and
  Zisserman}]{albanie2020bsl}
Samuel Albanie, G{\"u}l Varol, Liliane Momeni, Triantafyllos Afouras, Joon~Son
  Chung, Neil Fox, and Andrew Zisserman. 2020.
\newblock {BSL-1K}: {S}caling up co-articulated sign language recognition using
  mouthing cues.
\newblock In \emph{European Conference on Computer Vision (ECCV)}. Springer.

\bibitem[{Bull et~al.(2020)Bull, Gouiff{\`e}s, and
  Braffort}]{bull2020automatic}
Hannah Bull, Mich{\`e}le Gouiff{\`e}s, and Annelies Braffort. 2020.
\newblock Automatic segmentation of sign language into subtitle-units.
\newblock In \emph{European Conference on Computer Vision}, pages 186--198.
  Springer.

\bibitem[{Camgoz et~al.(2018)Camgoz, Hadfield, Koller, Ney, and
  Bowden}]{Phoenix2_dataset}
Necati~Cihan Camgoz, Simon Hadfield, Oscar Koller, Hermann Ney, and Richard
  Bowden. 2018.
\newblock \href {https://doi.org/10.1109/CVPR.2018.00812} {Neural sign language
  translation}.
\newblock In \emph{2018 IEEE/CVF Conference on Computer Vision and Pattern
  Recognition}, pages 7784--7793.

\bibitem[{{Cao} et~al.(2019){Cao}, {Hidalgo Martinez}, {Simon}, {Wei}, and
  {Sheikh}}]{OpenPose}
Z.~{Cao}, G.~{Hidalgo Martinez}, T.~{Simon}, S.~{Wei}, and Y.~A. {Sheikh}.
  2019.
\newblock Openpose: Realtime multi-person 2d pose estimation using part
  affinity fields.
\newblock \emph{IEEE Transactions on Pattern Analysis and Machine
  Intelligence}.

\bibitem[{Duarte et~al.(2022)Duarte, Albanie, Gir{\'o}-i Nieto, and
  Varol}]{duarte2022sign}
Amanda Duarte, Samuel Albanie, Xavier Gir{\'o}-i Nieto, and G{\"u}l Varol.
  2022.
\newblock Sign language video retrieval with free-form textual queries.
\newblock In \emph{Proceedings of the IEEE/CVF Conference on Computer Vision
  and Pattern Recognition}, pages 14094--14104.

\bibitem[{Duarte et~al.(2021)Duarte, Palaskar, Ventura, Ghadiyaram, DeHaan,
  Metze, Torres, and Giro-i Nieto}]{How2Sign_dataset}
Amanda Duarte, Shruti Palaskar, Lucas Ventura, Deepti Ghadiyaram, Kenneth
  DeHaan, Florian Metze, Jordi Torres, and Xavier Giro-i Nieto. 2021.
\newblock {How2Sign: A Large-scale Multimodal Dataset for Continuous American
  Sign Language}.
\newblock In \emph{Conference on Computer Vision and Pattern Recognition
  (CVPR)}.

\bibitem[{Gowda and May(2020)}]{greater_vocab_size}
Thamme Gowda and Jonathan May. 2020.
\newblock \href {https://doi.org/10.18653/v1/2020.findings-emnlp.352} {Finding
  the optimal vocabulary size for neural machine translation}.
\newblock In \emph{Findings of the Association for Computational Linguistics:
  EMNLP 2020}, pages 3955--3964, Online. Association for Computational
  Linguistics.

\bibitem[{Goyal et~al.(2022)Goyal, Gao, Chaudhary, Chen, Wenzek, Ju, Krishnan,
  Ranzato, Guzm{\'a}n, and Fan}]{low_resource_languages}
Naman Goyal, Cynthia Gao, Vishrav Chaudhary, Peng-Jen Chen, Guillaume Wenzek,
  Da~Ju, Sanjana Krishnan, Marc{'}Aurelio Ranzato, Francisco Guzm{\'a}n, and
  Angela Fan. 2022.
\newblock \href {https://doi.org/10.1162/tacl_a_00474} {The {F}lores-101
  evaluation benchmark for low-resource and multilingual machine translation}.
\newblock \emph{Transactions of the Association for Computational Linguistics},
  10:522--538.

\bibitem[{Hieber et~al.(2017)Hieber, Domhan, Denkowski, Vilar, Sokolov,
  Clifton, and Post}]{sockeye}
Felix Hieber, Tobias Domhan, Michael Denkowski, David Vilar, Artem Sokolov, Ann
  Clifton, and Matt Post. 2017.
\newblock \href {https://doi.org/10.48550/ARXIV.1712.05690} {Sockeye: A toolkit
  for neural machine translation}.

\bibitem[{Jiang et~al.(2021)Jiang, Sun, Wang, Bai, Li, and
  Fu}]{skeleton_aware_slr}
Songyao Jiang, Bin Sun, Lichen Wang, Yue Bai, Kunpeng Li, and Yun Fu. 2021.
\newblock Skeleton aware multi-modal sign language recognition.
\newblock In \emph{Proceedings of the IEEE/CVF Conference on Computer Vision
  and Pattern Recognition (CVPR) Workshops}.

\bibitem[{Kudo and Richardson(2018)}]{sentencepiece}
Taku Kudo and John Richardson. 2018.
\newblock \href {https://doi.org/10.18653/v1/D18-2012} {{S}entence{P}iece: A
  simple and language independent subword tokenizer and detokenizer for neural
  text processing}.
\newblock In \emph{Proceedings of the 2018 Conference on Empirical Methods in
  Natural Language Processing: System Demonstrations}, pages 66--71, Brussels,
  Belgium. Association for Computational Linguistics.

\bibitem[{Li et~al.(2020)Li, Rodriguez, Yu, and Li}]{li2020wlasl}
Dongxu Li, Cristian Rodriguez, Xin Yu, and Hongdong Li. 2020.
\newblock Word-level deep sign language recognition from video: A new
  large-scale dataset and methods comparison.
\newblock In \emph{Proceedings of the IEEE/CVF winter conference on
  applications of computer vision}, pages 1459--1469.

\bibitem[{Lugaresi et~al.(2019)Lugaresi, Tang, Nash, McClanahan, Uboweja, Hays,
  Zhang, Chang, Yong, Lee et~al.}]{MediaPipe}
Camillo Lugaresi, Jiuqiang Tang, Hadon Nash, Chris McClanahan, Esha Uboweja,
  Michael Hays, Fan Zhang, Chuo-Ling Chang, Ming~Guang Yong, Juhyun Lee, et~al.
  2019.
\newblock {Mediapipe: A framework for building perception pipelines}.
\newblock \emph{ArXiv}, abs/1906.08172.

\bibitem[{Moryossef(2022)}]{pose_format_helper}
Amit Moryossef. 2022.
\newblock Complete toolkit for working with poses.
\newblock \url{https://github.com/AmitMY/pose-format/}.

\bibitem[{M\"{u}ller et~al.(2022{\natexlab{a}})M\"{u}ller, Ebling, Camgöz,
  Jian, Battisti, Tissi, Sidler-Miserez, Perrollaz, Berger, Reinhard,
  Moryossef, Rios, Bowden, Wong, Ribback, and Schori}]{srf}
Mathias M\"{u}ller, Sarah Ebling, Necati~Cihan Camgöz, Zifan Jian, Alessia
  Battisti, Katja Tissi, Sandra Sidler-Miserez, Regula Perrollaz, Michèle
  Berger, Sabine Reinhard, Amit Bar-Ilan~University Moryossef, Annette Rios,
  Richard Bowden, Ryan Wong, Robin Ribback, and Severine Schori.
  2022{\natexlab{a}}.
\newblock \href {https://doi.org/10.5281/zenodo.6637392} {{WMT-SLT SRF:
  Training data for the WMT shared task on sign language translation (videos,
  subtitles)}}.
\newblock {We additionally acknowledge funding through the Innosuisse Flagship
  "Inclusive Information and Communication Technologies" (IICT) (grant
  agreement no. PFFS-21-47).}

\bibitem[{M\"{u}ller et~al.(2022{\natexlab{b}})M\"{u}ller, Ebling, Camgöz,
  Jiang, Battisti, Moryossef, Rios, Bowden, and Wong}]{focusnews}
Mathias M\"{u}ller, Sarah Ebling, Necati~Cihan Camgöz, Zifan Jiang, Alessia
  Battisti, Amit Moryossef, Annette Rios, Richard Bowden, and Ryan Wong.
  2022{\natexlab{b}}.
\newblock \href {https://doi.org/10.5281/zenodo.6631159} {{WMT-SLT FocusNews:
  Training data for the WMT shared task on sign language translation}}.
\newblock {We additionally acknowledge funding through the Innosuisse Flagship
  "Inclusive Information and Communication Technologies" (IICT) (grant
  agreement no. PFFS-21-47).}

\bibitem[{M{\"u}ller et~al.(2022)M{\"u}ller, Ebling, Eleftherios, Battisti,
  Berger, Bowden, Braffort, Camg{\"o}z, Espa{\~n}a-Bonet, Grundkiewicz, Jiang,
  Koller, Moryossef, Perrollaz, Reinhard, Rios, Shterionov, Sidler-Miserez,
  Tissi, and Van~Landuyt}]{wmt_slt_preliminary_findings}
Mathias M{\"u}ller, Sarah Ebling, Avramidis Eleftherios, Alessia Battisti,
  Mich{\`e}le Berger, Richard Bowden, Annelies Braffort, Necati~Cihan
  Camg{\"o}z, Cristina Espa{\~n}a-Bonet, Roman Grundkiewicz, Zifan Jiang, Oscar
  Koller, Amit Moryossef, Regula Perrollaz, Sabine Reinhard, Annette Rios,
  Dimitar Shterionov, Sandra Sidler-Miserez, Katja Tissi, and Davy Van~Landuyt.
  2022.
\newblock Findings of the {WMT} 2022 shared task on sign language translation.
\newblock In \emph{Proceedings of the Seventh Conference on Machine
  Translation}, Abu Dhabi, United Arab Emirates. Association for Computational
  Linguistics.

\bibitem[{M\"{u}ller et~al.(2022)M\"{u}ller, Rios, and
  Moryossef}]{mueller2022sign-sockeye-baselines}
Mathias M\"{u}ller, Annette Rios, and Amit Moryossef. 2022.
\newblock Sockeye baseline models for sign language translation.
\newblock \url{https://github.com/bricksdont/sign-sockeye-baselines}.

\bibitem[{Ott et~al.(2019)Ott, Edunov, Baevski, Fan, Gross, Ng, Grangier, and
  Auli}]{Fairseq}
Myle Ott, Sergey Edunov, Alexei Baevski, Angela Fan, Sam Gross, Nathan Ng,
  David Grangier, and Michael Auli. 2019.
\newblock fairseq: A fast, extensible toolkit for sequence modeling.
\newblock In \emph{Proceedings of NAACL-HLT 2019: Demonstrations}.

\bibitem[{Papineni et~al.(2002)Papineni, Roukos, Ward, and Zhu}]{bleu}
Kishore Papineni, Salim Roukos, Todd Ward, and Wei-Jing Zhu. 2002.
\newblock \href {https://doi.org/10.3115/1073083.1073135} {{B}leu: a method for
  automatic evaluation of machine translation}.
\newblock In \emph{Proceedings of the 40th Annual Meeting of the Association
  for Computational Linguistics}, pages 311--318, Philadelphia, Pennsylvania,
  USA. Association for Computational Linguistics.

\bibitem[{Popel and Bojar(2018)}]{training_tips_transformer}
Martin Popel and Ond{\v{r}}ej Bojar. 2018.
\newblock Training tips for the transformer model.
\newblock \emph{The Prague Bulletin of Mathematical Linguistics}, 110:43--70.

\bibitem[{Popovi{\'c}(2017)}]{chrf++}
Maja Popovi{\'c}. 2017.
\newblock \href {https://doi.org/10.18653/v1/W17-4770} {chr{F}++: words helping
  character n-grams}.
\newblock In \emph{Proceedings of the Second Conference on Machine
  Translation}, pages 612--618, Copenhagen, Denmark. Association for
  Computational Linguistics.

\bibitem[{Salesky et~al.(2020)Salesky, Runge, Coda, Niehues, and
  Neubig}]{choosing_vocab_size}
Elizabeth Salesky, Andrew Runge, Alex Coda, Jan Niehues, and Graham Neubig.
  2020.
\newblock \href {https://doi.org/10.1007/s10590-019-09243-8} {Optimizing
  segmentation granularity for neural machine translation}.
\newblock \emph{Machine Translation}, 34(1):41–59.

\bibitem[{Sellam et~al.(2020)Sellam, Das, and Parikh}]{bleurt}
Thibault Sellam, Dipanjan Das, and Ankur Parikh. 2020.
\newblock \href {https://doi.org/10.18653/v1/2020.acl-main.704} {{BLEURT}:
  Learning robust metrics for text generation}.
\newblock In \emph{Proceedings of the 58th Annual Meeting of the Association
  for Computational Linguistics}, pages 7881--7892, Online. Association for
  Computational Linguistics.

\bibitem[{Sennrich et~al.(2016)Sennrich, Haddow, and Birch}]{bpe_tokenizer}
Rico Sennrich, Barry Haddow, and Alexandra Birch. 2016.
\newblock \href {https://doi.org/10.18653/v1/P16-1162} {Neural machine
  translation of rare words with subword units}.
\newblock In \emph{Proceedings of the 54th Annual Meeting of the Association
  for Computational Linguistics (Volume 1: Long Papers)}, pages 1715--1725,
  Berlin, Germany. Association for Computational Linguistics.

\bibitem[{Tran et~al.(2021)Tran, Bhosale, Cross, Koehn, Edunov, and
  Fan}]{sota_mt_news}
Chau Tran, Shruti Bhosale, James Cross, Philipp Koehn, Sergey Edunov, and
  Angela Fan. 2021.
\newblock \href {https://aclanthology.org/2021.wmt-1.19} {{F}acebook {AI}{'}s
  {WMT}21 news translation task submission}.
\newblock In \emph{Proceedings of the Sixth Conference on Machine Translation},
  pages 205--215, Online. Association for Computational Linguistics.

\bibitem[{Vaezi~Joze and Koller(2019)}]{joze2018msasl}
Hamid Vaezi~Joze and Oscar Koller. 2019.
\newblock Ms-asl: A large-scale data set and benchmark for understanding
  american sign language.
\newblock \emph{The British Machine Vision Conference (BMVC)}.

\bibitem[{Vaswani et~al.(2017)Vaswani, Shazeer, Parmar, Uszkoreit, Jones,
  Gomez, Kaiser, and Polosukhin}]{attention_is_all_you_need}
Ashish Vaswani, Noam Shazeer, Niki Parmar, Jakob Uszkoreit, Llion Jones,
  Aidan~N Gomez, \L~ukasz Kaiser, and Illia Polosukhin. 2017.
\newblock \href
  {https://proceedings.neurips.cc/paper/2017/file/3f5ee243547dee91fbd053c1c4a845aa-Paper.pdf}
  {Attention is all you need}.
\newblock In \emph{Advances in Neural Information Processing Systems},
  volume~30. Curran Associates, Inc.

\bibitem[{Voskou et~al.(2021)Voskou, Panousis, Kosmopoulos, Metaxas, and
  Chatzis}]{stochastic_transformers_phoenix}
Andreas Voskou, Konstantinos~P. Panousis, Dimitrios Kosmopoulos, Dimitris~N.
  Metaxas, and Sotirios Chatzis. 2021.
\newblock Stochastic transformer networks with linear competing units:
  Application to end-to-end sl translation.

\bibitem[{Yan et~al.(2019)Yan, Xiong, Jingbo, and Lin}]{mmskeleton2019}
Sijie Yan, Yuanjun Xiong, Wangm Jingbo, and Dahua Lin. 2019.
\newblock Mmskeleton.
\newblock \url{https://github.com/open-mmlab/mmskeleton}.

\bibitem[{Yan et~al.(2018)Yan, Xiong, and Lin}]{stgcn2018aaai}
Sijie Yan, Yuanjun Xiong, and Dahua Lin. 2018.
\newblock Spatial temporal graph convolutional networks for skeleton-based
  action recognition.
\newblock In \emph{AAAI}.

\bibitem[{Yang et~al.(2021)Yang, Ma, Huang, Zhang, Dong, Huang, Muzio, Singhal,
  Hassan, Song, and Wei}]{sota_mt_multi}
Jian Yang, Shuming Ma, Haoyang Huang, Dongdong Zhang, Li~Dong, Shaohan Huang,
  Alexandre Muzio, Saksham Singhal, Hany Hassan, Xia Song, and Furu Wei. 2021.
\newblock \href {https://aclanthology.org/2021.wmt-1.54} {Multilingual machine
  translation systems from {M}icrosoft for {WMT}21 shared task}.
\newblock In \emph{Proceedings of the Sixth Conference on Machine Translation},
  pages 446--455, Online. Association for Computational Linguistics.

\end{thebibliography}
\bibliographystyle{acl_natbib}

\end{document}